\documentclass{article}

\usepackage{amsmath,amsfonts,bm}

\def\eqref#1{equation~\ref{#1}}

\def\1{\bm{1}}

\DeclareMathAlphabet{\mathsfit}{\encodingdefault}{\sfdefault}{m}{sl}
\SetMathAlphabet{\mathsfit}{bold}{\encodingdefault}{\sfdefault}{bx}{n}

\usepackage{amsmath,amsfonts,bm}
\usepackage{microtype}
\usepackage{graphicx}
\usepackage{subfigure}
\usepackage{booktabs} 
\usepackage{amsmath}
\usepackage{multirow}
\usepackage{amssymb}
\usepackage{diagbox}
\usepackage{enumitem}
\usepackage{url}
\usepackage{amsfonts}       
\usepackage{nicefrac}       
\usepackage{microtype}      
\usepackage{graphicx}
\usepackage{wrapfig}
\usepackage{xcolor}
\usepackage{multirow}
\usepackage{algorithm}
\usepackage{algorithmic}
\usepackage{eqparbox}
\usepackage{multirow}
\usepackage{float}
\usepackage{setspace}
\usepackage{tikz}
\usepackage{hyperref}
\usepackage[accepted]{icml2020}

\newcommand{\wteacher}{w}
\newcommand{\wstudent}{\hat{w}}

\icmltitlerunning{Learning to Remember from a Multi-Task Teacher}

\begin{document}
\twocolumn[
\icmltitle{Learning to Remember from a Multi-Task Teacher}
\icmlsetsymbol{equal}{*}
\icmlsetsymbol{uber}{3}

\begin{icmlauthorlist}
\icmlauthor{Yuwen Xiong}{equal}
\icmlauthor{Mengye Ren}{equal}
\icmlauthor{Raquel Urtasun}{}
\end{icmlauthorlist}

\icmlcorrespondingauthor{Yuwen Xiong}{yuwen@uber.com}
\icmlcorrespondingauthor{Mengye Ren}{mren3@uber.com}
\icmlcorrespondingauthor{Raquel Urtasun}{urtasun@uber.com}
\icmlkeywords{meta-learning, catastrophic forgetting, multi-task learning}
\vskip 0.3in
]

\printAffiliationsAndNotice{} 
\vspace{-0.2cm}
\begin{abstract}
Recent studies on catastrophic forgetting during sequential learning typically focus on fixing the
accuracy of the predictions for a previously learned task. In this paper we argue that the outputs
of neural networks are subject to rapid changes when learning a new data distribution, and
networks that appear to ``forget'' everything still contain useful representation towards previous
tasks. Instead of enforcing the output accuracy to stay the same, we propose to reduce the effect
of catastrophic forgetting on the representation level, as the output layer can be quickly
recovered later with a small number of examples. Towards this goal, we propose an experimental
setup that measures the amount of representational forgetting, and  develop a novel meta-learning
algorithm to overcome this issue. The proposed meta-learner produces weight updates of a
sequential learning network, mimicking a multi-task teacher network's representation. We show that
our meta-learner can improve its learned representations on new tasks, while maintaining a good
representation for old tasks.
\end{abstract}

\vspace{-0.3in}
\section{Introduction}

An intelligent agent  needs to deal with a dynamic world and is typically presented with sequential
tasks that are highly correlated in time yet constantly changing. Newborns learn to build generic
representations from video and audio streaming input. Kids can learn highly skilled tasks such as
skiing and swimming sequentially without worrying about forgetting one another. Humans seem to have
a robust way of learning representations from sequential inputs (and tasks), yet  state-of-the-art
machine learning algorithms rely heavily on uniformly sampled training examples from the same
distribution.

One of the major challenges in sequential learning of neural networks is the issue of {\it
catastrophic forgetting} \citep{mccloskey1989catastrophic}--after a neural network is trained on a
new task, its performance on old tasks drops significantly. Despite several attempts, this problem
remains unsolved. Explicit weight regularization methods \citep{l2,ewc,imm} often rely on simplistic
assumptions on the shape of the weight posterior distribution. Model compression methods
\citep{hardattend,pathnet,packnet} seem promising on existing benchmarks, however the underlying
mechanism is to train small individual networks that may lack global cooperation, a limiting factor
when learning a large number of classes towards a generic representation. Generative models
\citep{dgr,fearnet,uncompromising} seem to be a natural choice; however, training a high quality
generative model is far from trivial, oftentimes  more complex than training the original network
itself.

Despite the variety of models that have been proposed, there seems to be a lack of general
understanding on what kind of knowledge is being forgotten and to what extent it can be recovered.
Recent research places much of its focus on maintaining the output performance of previous tasks. In
this paper we argue that this  can be misleading since the output layer of a network is very
sensitive to changes in the output distribution. Instead, here we would like to understand how much
of the performance drop is related to the lack of training on previous output layers versus the loss
of information in the newly learned representation. Towards this goal, we exploit a linear
decoding layer to measure the amount of catastrophic forgetting on the representation level. This
gives us insights on whether the drop in performance is likely to be recovered by re-learning the
output layer from very few examples.

Motivated by recent progress on meta-learning \citep{syngrad,l2l,learnunsup}, in this paper we
propose to learn a weight update rule to overcome representational forgetting. our meta-learning
algorithm learn such a rule by rolling out many sequential learning experiences during training. In
human language acquisition, it is found that children who lost their first language maintain similar
brain activation to bilingual speakers~\citep{lostfirstlang}. Inspired by this fact, we propose a
novel meta-learning algorithm that tries to mimic a multi-task teacher network's representation, an
offline oracle in our sequential learning setup, since multi-task learning has simultaneous access
to all tasks whereas our sequential learning algorithm only has access to one task at a time. 

In summary, the contributions of our paper are two-fold. First, we propose a measure of catastrophic
forgetting at the representation level, which provides more insight on the amount of previous
knowledge forgotten in the new task. Second, we develop a new meta-learning algorithm that can
predict weight updates that are less prone to catastrophic forgetting than standard backpropagation.
We demonstrate the effectiveness of our approach on the CIFAR-10, CIFAR-100~\citep{cifar} and Tiny-ImageNet\footnote{\url{https://tiny-imagenet.herokuapp.com/}} datasets, and find that our
meta-learner is able to generalize to unseen object classes from meta-training.

The rest of the paper is organized as follows: we first survey existing literature in
Section~\ref{sec:related}; Section~\ref{sec:readout} describes representational forgetting in deep
neural networks, and our experimental setup to measure it; Section~\ref{sec:alg} details our
proposed meta-learning algorithm, followed by experimental results in
Section~\ref{sec:exp}.
 \vspace{-0.1in}
\section{Related Work}
\label{sec:related}

Catastrophic forgetting in sequential learning has been studied in the early literature of neural
networks~\citep{mccloskey1989catastrophic,ratcliff1990connectionist,french1991semi,pretrain,french1999catastrophic}.
The degree of forgetting is usually measured in terms of the amount of time, typically the number of
iterations, saved to relearn the old task. This metric has several drawbacks as the number of
iterations can be fairly sensitive to the choice of optimization hyper-parameters and network
architecture. Furthermore, the network may never recover fully the original performance.

With the recent success of deep learning, the issue of catastrophic forgetting has regained
attention in the research community. Unlike classical methods, recent papers measure the old task
performance immediately after training on the new task~\citep{ewc,empirical,lwf}. In this setting,
networks that are trained in a sequential manner over tasks suffer dramatically, due to the fact
that the previous output classifier branches are no longer tuned to the newly learned representations.

In this paper we follow the early literature, and allow the network to relearn on old tasks, as we
believe it is natural to review the old task before testing on it again. But instead of counting the
number of iterations to recover the old task performance, we propose to train on top of the newly
learned representation using a small decoding model to perform the old task. We believe this measure
well captures the amount of representational forgetting. To this end, we exploit a simple linear
readout layer, as it is relatively fast to train and is more robust than measuring the number of
recovering steps.

One way to address catastrophic forgetting is through explicit regularization. \citet{l2} add an L2
regularizer to ensure that the new weights do not ``drift'' away from the old weights. Elastic
weight consolidation~\citep{ewc} computes the strength of the regularizer on each weight dimension
using a diagonal approximation of the Fisher information matrix. \citet{synapticintell} propose to
directly approximate the regularization strength online. \citet{multimodel} uses similar
regularization function to counter interference in multi-task learning. \citet{imm} incrementally
match the moments of posterior Gaussian distributions of new tasks. These methods are often
motivated from a simple quadratic loss surface, and can be potentially limiting their ability to
deal with more complex learning dynamics. Regularization can also be imposed on the activation
level: Learning without Forgetting~\citep{lwf} regularizes the network such that the logits of the
new examples on the old classifier remain similar. While this framework is more flexible, using the
old activations to distill new tasks can  be less informative if the network has not seen enough
classes.

In contrast to continuous regularization, model compression based
approaches~\citep{packnet,pathnet,pnn,expandable} discretely allocate certain capacity of a network
towards learning new tasks. PackNet~\citep{packnet} applies network pruning in between sequential
tasks, so that the pruned neurons can be re-allocated. PathNet~\citep{pathnet} uses genetic
algorithms to select pathways of the network for reuse. HAT~\citep{hardattend} learns a different
task-oriented hard binary mask for each hidden unit. \citet{pnn} add connections from old frozen
modules towards newly allocated modules, at the price of learning more intermediate layers, thus
scaling quadratically with the number of tasks. \citet{expandable} propose to dynamically prune and
allocate neurons at the same time. In comparison, our meta-learner can also be interpreted as
implicitly learning to perform dynamic capacity re-allocation without increasing the network size;
however, our approach does not choose a discrete set of neurons or synapses to update, but try to
mimic the activation responses from a multi-task network. This allows the algorithm to learn to
implicitly allocate weight subspaces for learning the new tasks.

Another class of methods store a subset of the old data, so that the old task can be jointly
trained. iCaRL~\citet{icarl} propose to choose representative exemplars of old tasks. Gradient
episodic memory~\citep{gem} stores old examples and makes sure that the new example only updates in
the direction that agrees with the gradient directions of old examples. 
\citet{mbpa} explore a learnable memory architecture to dynamically store and
retrieve examples facilitating sequential learning. Though effective, storing raw data points costs additional storage and it may also lack biological plausibility.
To address the issue of data storage, generative models have also been used to avoid
storing old raw data~\citep{dgr,uncompromising,fearnet,generativereplay}. 
Although generative models enjoy the
benefits of data storage based models, 
the final performance heavily depends on the quality of the
generated data, because training a competitive generative model itself may be more complex and take more
capacity than the original network.

Our proposed model is inspired from prior work in  meta-learning: using a learned parameterized
weight update rule~\citep{bengio1990learning} instead of standard optimization methods. Synthetic
gradient~\citep{syngrad} uses an MLP to predict the gradient direction when performing forward
passes, allowing asynchronous weight updates across layers. \citet{l2l,metalstm} use a recurrent
network to predict updates. \citet{learnunsup} propose to learn an unsupervised learning rule based
solely on activations. \citet{diffplasticity,backpropamine} combine a learned Hebbian plasticity
rule with learned weights. The largest difference between our proposed model and prior work is the
fact that instead of predicting the gradients or improve the task performance at the end of the
training episode, our meta-learner is trained with a teacher network's activation as supervision.
\vspace{-0.1in}
\section{Representational Forgetting in Sequential Learning}
\label{sec:readout}
\begin{figure}[t]
\centering
\vspace{-0.1in}
\includegraphics[width=\columnwidth,trim={1.5cm 10cm 15.8cm 0.1cm},clip]{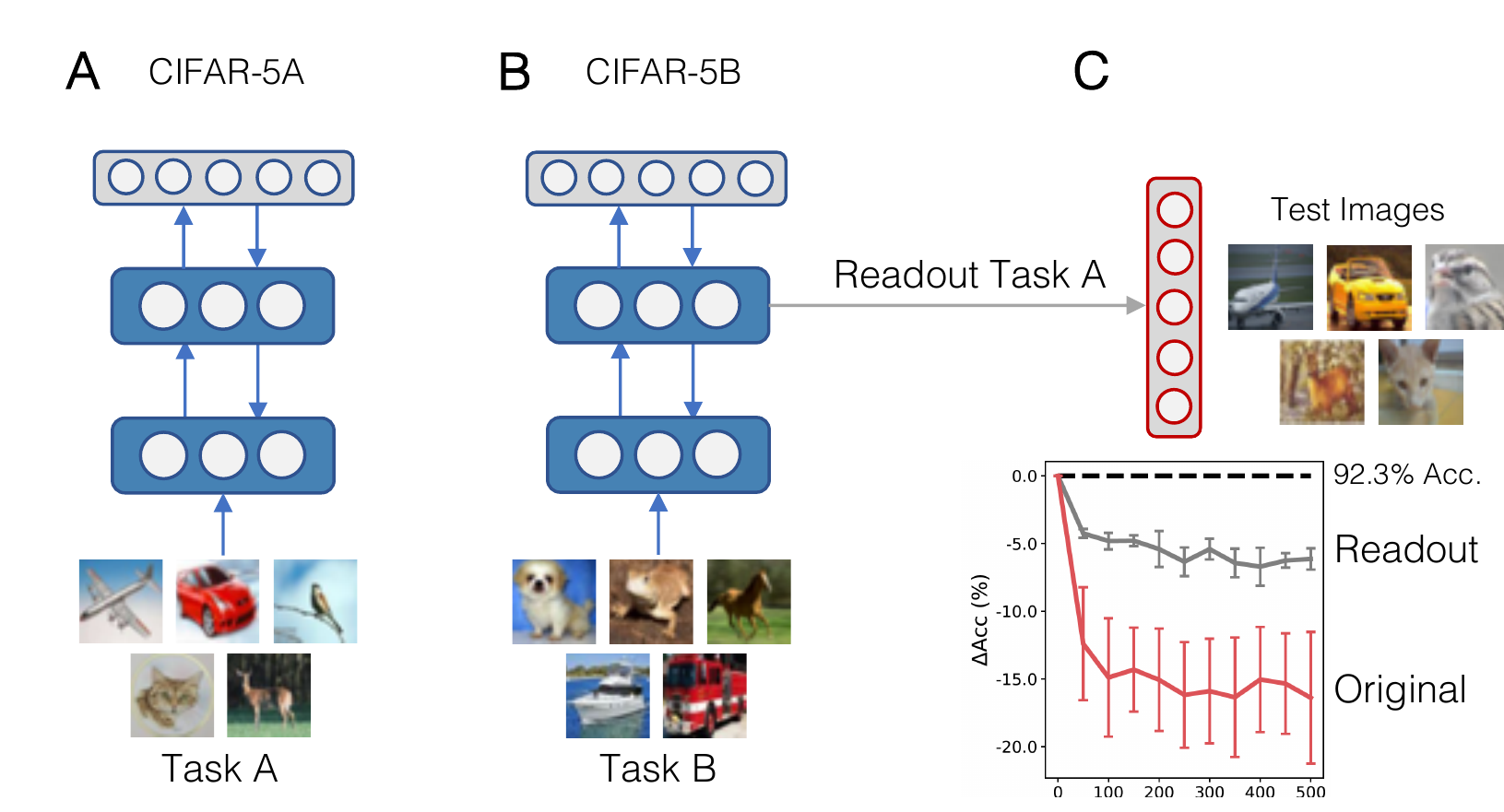}
\vspace{-0.1in}
\caption{Using a readout layer to measure catastrophic forgetting on representations. \textbf{A)} A
network is first pre-trained on Task A. \textbf{B)} Then it is finetuned on Task B. \textbf{C)} Task A training data is then fed to the network, and the representation 
prior to the classification layer is recorded. We re-train a readout layer to recover Task A output. Test
accuracy on Task A is ``Readout'', and original classification layer accuracy is
``Original''. Chance is 20\%.}
\label{fig:continual}
 \vspace{-0.25in}
\end{figure}

Sequential learning is the process of learning tasks sequentially without revisiting previous tasks.
At each incremental learning stage, the network has learned a set of source tasks and then is
presented with a novel target task. For simplicity, we refer the union of all source tasks as
\textit{Task A}, and the novel task as \textit{Task B}.

The exposure to Task A can bring positive benefits
towards learning Task B if the two tasks are similar. This is a property  often studied in the
transfer learning literature. Unfortunately, it is challenging to maintain the initial  performance
of the model learned on Task A (when updating it to perform well on task B), especially when the old
task environment is not available to the agent. For example, a commercial robot needs to adapt and
learn in new environments while the original training data cannot be shipped together with the
learning algorithm.

Catastrophic forgetting~\citep{mccloskey1989catastrophic} occurs when a learning agent forgets the
old task after adapting to the new task. While recent literature solely focuses on the output
performance of an agent on the previous tasks, it is unclear whether the agent ``truly forgets'' the
prior experiences, or only the output layers are miscalibrated due to learning the new task. The
latter issue has an easy solution as simply training the output classifier on the old task for a few
iterations can recover most of the performance loss. This is similar to human revisiting previously
learned skills.

In this paper we study how much forgetting occurs at the representation level. Towards this goal, we
propose the experiment setup illustrated in Figure~\ref{fig:continual} where a network is first
trained on Task A, and it learns the representation of the inputs, referred to as the final layer
prior to the classification head. Next, a different task, Task B is introduced. In the second stage of
training, the model no longer has access to Task A. It then fine-tunes all its layers on Task B,
since learning a better representation will be useful for the new task. To test the amount of
forgetting on Task A, we train a linear readout layer on the newly learned representations, using the
training examples from Task A, and evaluate the test performance.

\begin{figure*}[t]
\vspace{-0.1in}
\centering
\includegraphics[width=0.98\textwidth,trim={0cm 9.3cm 0cm 0.5cm},clip]{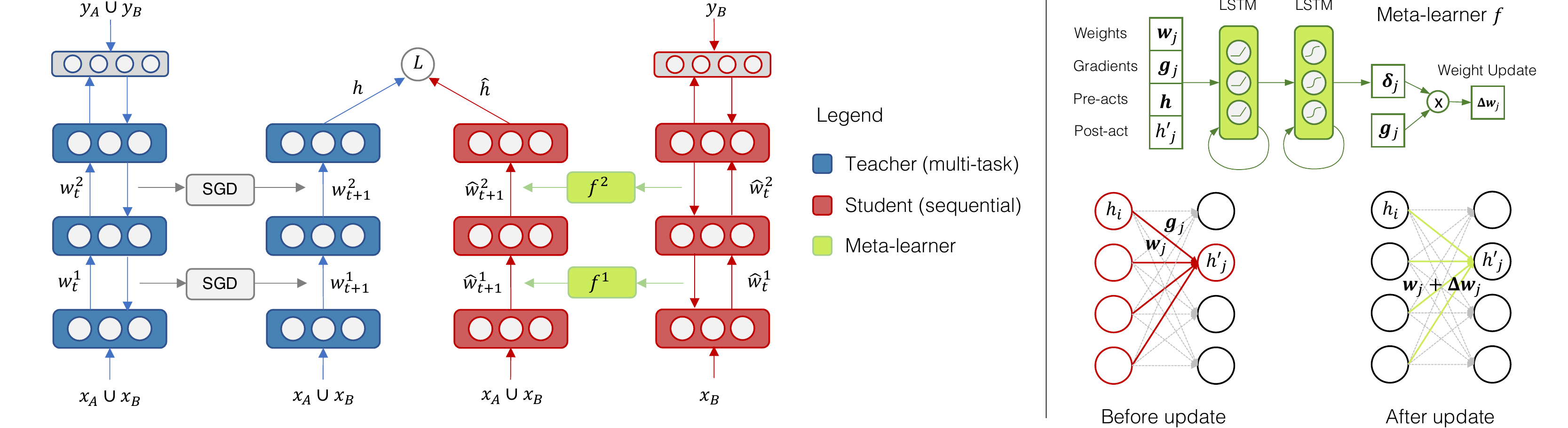}
\vspace{-0.05in}
\caption{
\textbf{Left:} Overview of our purposed method during one training step: 1) the teacher network is updated
using SGD on multi-task data; 2) the student network is updated using the meta-learner on Task B
data only; 3) the multi-task data are fed into both networks and we record the representations
as $h$ and $\hat{h}$; 4) the meta module is then updated to minimize the difference between $h$ and
$\hat{h}$. At meta-test time, the teacher network is no longer present, and we update the student
network solely with the trained meta-learner for the entire training sequence.
\textbf{Right:} Our meta-learner module is a stacked LSTM network. For a given output neuron $h'_j$, the
meta-network takes in its activation, its pre-activations, current weights and gradients, and output
the weight updates associated with $h'_j$. Weights in the same layer will be shared a single
meta-network,  while different layers will have different meta-networks.
}
\label{fig:main}
\vspace{-0.1in}
\end{figure*}

We repeatedly do this readout training throughout the learning of task B. In the beginning, the
readout performance is close to the original Task A performance, and later, the features become less
selective towards Task A. As shown in Figure~\ref{fig:continual}, we indeed observe that the readout
accuracy of Task A is constantly decreasing as the learning process goes on. In contrast to
measuring the output performance directly (``Readout'' vs. ``Original'' in
Figure~\ref{fig:continual}), as was done in prior catastrophic forgetting literature, the level of
forgetting here is not as dramatic, suggesting a portion of forgetting happens due to the
miscalibration of the output layer. Representational forgetting is still significant--for the
binary classification problem illustrated in Figure~\ref{fig:continual}, readout accuracy on Task A
drops over 15\% after 300 iterations of training on Task B.

Meta-learning is a general tool for us to learn a new learning algorithm with desired properties. In
the next section, we propose a novel meta-learning algorithm that directly addresses the issue of
representational forgetting.

\vspace{-0.1in}
\section{Learning from a Multi-Task Teacher}
\label{sec:alg}

As the learning process goes on in Task B, each gradient descent step can potentially erase useful
features for the old task. Continuous regularization~\citep{l2,ewc,synapticintell} or discrete model
pruning based methods~\citep{packnet,pathnet,pnn,expandable}, as discussed in
Section~\ref{sec:related}, can hurt the capacity of learning new tasks and limit the sharing of a
distributed representation. Manually designing a sophisticated learning objective is not obvious,
thus we are interested in learning a \textit{learning rule} that can predict dynamic weight updates
alongside the training of Task B. 

For simplicity, we describe the setting of a fully connected layer, shown in Figure~\ref{fig:main},
where $i$ indexes pre-activations and $j$ post-activations. A weight synapse $w_{i,j}$ connects to
its pre-activation $h_i$ and post-activation $h'_j$. Let $g_{i,j}$ be the gradient of the connection
obtained through regular backpropagation. Our meta-learner $f$ is implemented as a long short-term
memory (LSTM) network that takes as inputs the weight connection, gradients, inputs and outputs,
similar to what has been done by \cite{learnunsup}. The network intuitively predicts a dynamic
gating that modulates the plasticity of synapses: $\delta_{\cdot,j} = f(\mathbf{w}_{\cdot,j},
\mathbf{g}_{\cdot,j}, \mathbf{h}_{\cdot}, h'_j; \theta)$, which is then multiplied with the original
gradients to form the updates of the synapse: $\Delta w_{\cdot,j} = \delta_{\cdot,j} \cdot
\mathbf{g}_{\cdot,j}$. Finally, the weights are updated in a similar way to SGD,
$\hat{\mathbf{w}}_{\cdot,j} = \mathbf{w}_{\cdot,j} -\alpha \Delta
\mathbf{w}_{\cdot,j}$, where $\alpha$ is the learning rate. We can generalize this setup to
convolutional layers as well by considering all filter locations together, with more details in
Section~\ref{sec:impl}.

Learning this meta-learner, is however a non-trivial task. In \citep{learnunsup},
the learning process is unrolled just like a recurrent neural network for a large number of steps,
and meta-learning is done using backpropagation through time, which is inefficient since each
meta-update can only be done in the outer loop.

A key insight of our paper is that, we can actually learn from a multi-task teacher network as if
it is an oracle to our sequential learner. Similar to knowledge distillation~\citep{distillation},
where a student network learns activation patterns from a teacher network, here, we would like to
train the meta-learner such that the student network has a similar activation pattern compared to
the teacher network. This avoids using the final accuracy of an episode as the supervision signal,
which can be very inefficient.

The idea of using a multi-task network as a teacher is inspired by the human language acquisition
literature. It was found that adopted children who were separated from the Chinese language (their
birth language) at around one year old on average, still maintain activations in their left superior
temporal gyrus similar to French/Chinese bilingual speakers at the age of 9 to 17, even though they
have no exposure to Chinese for 13 years on average~\citep{lostfirstlang}. We draw a parallel in our
sequential learning framework here: whereas the network is allowed to forget the old task on the
output level, the learned representation should resemble the one learned by a multi-task network.

The overall algorithm has two nested loops, like many other meta-learning algorithms. The inner loop
simulates the experience of doing a single learning process, and the outer loop rolls the network
back to its initial point and re-starts the learning again. Different from standard hyper-parameter
optimization, in our proposed algorithm, the meta-learner is updated every step in the inner loop,
which makes the training more efficient. The meta-training procedure is detailed in
Algorithm~\ref{alg:main}.

In the beginning of each inner loop, the teacher network and the student network share the same
initialization that is pre-trained on Task A. For every step in the inner loop, as illustrated in
Figure~\ref{fig:main}, the teacher network performs a regular SGD update using data from both the
old and new tasks. The student network, who does not have access to the old task, computes the
gradients on the new task and sends them to the meta-learner network, which then predicts a
multiplicative gating. Now using the newly updated weights, we compare the representation difference
between the teacher and student networks, and update the meta-learner to minimize this difference.

\vspace{-0.01in}
\begin{algorithm}[H]
\begin{small}
\caption{Learning to Remember}
\label{alg:main}
\begin{algorithmic}[1]
\REQUIRE $w_0$, $\mathcal{D}_{1\dots K}$
\ENSURE $\theta$ (Meta-parameters)
\FOR{$i=1$ ... $N$}
\STATE $\wteacher \gets w_0$ \quad \COMMENT{Reinitialize teacher and student networks}
\STATE $\wstudent \gets w_0$
\STATE Meta-learner resets hidden state;
\STATE \COMMENT{Reinitialize T-BPTT step $s$ to zero}
\STATE $s \gets 0$ 
\FOR{$k=2 ... K$}

\STATE $\mathcal{D}_A \gets \mathcal{D}_{1 \dots k-1}$;
\STATE $\mathcal{D}_B \gets \mathcal{D}_{k}$ \quad \COMMENT{New Task};
\FOR{$t=1$ ... $T(i)$}
    \STATE $x_a, y_a \gets \text{Sample}(\mathcal{D}_A)$; $x_b, y_b \gets \text{Sample}(\mathcal{D}_B)$;
    \STATE $g \gets \text{TeacherNetBackward}(x_a \cup x_b, y_a \cup y_b; \wteacher)$;
    \STATE $\hat{g} \gets \text{StudentNetBackward}(x_b, y_b; \wstudent)$;
    \STATE \COMMENT{Teacher update}
    \STATE $w \gets w - \alpha g$;
    \STATE \COMMENT{Student update with meta-learner $f$}
    \STATE $\Delta \wstudent \gets \hat{g} \cdot f(\wstudent, \hat{g}, \hat{h};\theta)$;
    \STATE $\wstudent \gets \wstudent - \alpha \Delta \wstudent$; 
    \STATE $h \gets \text{TeacherNet}(x_a \cup x_b; w)$;
    \STATE $\hat{h} \gets \text{StudentNet}(x_a \cup x_b; \hat{w})$;
    \STATE $L \gets \text{HuberLoss}(h, \hat{h})$;
    \STATE $s \gets s+1$;
    \STATE \COMMENT{Meta-learner update, backprop thru time $s$ steps}
    \STATE $\theta \gets \theta - \eta $T-BPTT$(L, \theta, s)$;
    \IF{$L >$ LossThreshold($i$)}
        \label{alg:curriculum}
        \STATE Restart inner-loop learning
    \ENDIF
    \IF{$s \ge$ T-BPTTSteps($i$)}
        \label{alg:tbptt}
        \STATE $s \gets 0$; // Reset T-BPTT steps
    \ENDIF
\ENDFOR
\ENDFOR
\ENDFOR
\end{algorithmic}
\end{small}
\end{algorithm}
\vspace{-0.2in}

We use Huber loss multiplied with a scalar hyperparameter as the objective for minimizing the
representational differences. To help the meta-learner gradually make progress, we set up a
curriculum such that whenever the loss is greater than certain threshold, we will reinitialize the
learning process to prevent the meta-learner deviating too far (see Line~\ref{alg:curriculum} of
Algorithm~\ref{alg:main}). To speed up learning, we perform truncated backpropagation through time
on the meta-learner LSTM. Gradient accumulation is reset whenever the number of unrolled steps is
longer than the truncation steps (see Line~\ref{alg:tbptt} of Algorithm~\ref{alg:main}).

\vspace{-0.1in}
\section{Experiments}
\label{sec:exp}
In this section, we first give  implementation details of our algorithm, and then  report
experimental results on five sets of experiments, detailed in Table~\ref{tab:tasks}. In the first
set, we test a setting where meta-testing samples the same task distribution as meta-training. In
the second and third sets of experiments, we verify the generalization ability of our meta-learner,
by using unseen classes and unseen initialization checkpoints for evaluation. In the fourth sets, we
verify whether the algorithm can generalize to a new dataset; and lastly, we try longer sequences
with three sequential tasks. For all results in the figures, we report an average of 5 runs.

\begin{table*}[t]
 \vspace{-0.05in}
\begin{center}
\begin{small}
\resizebox{0.95\textwidth}{!}{
\begin{tabular}{clccccccccc}
\toprule
No. & Experiment                      & \# Tasks & \# Cls & \# Steps      & Unseen Cls  & Unseen Network & Unseen Domain & Source & Meta-Train  & Meta-Test                          \\
\midrule                                                                                                                                                                         
1& Seq. Transfer                & 2        & 10     &  500          &             &                &               & CIFAR-5A & CIFAR-5B  & CIFAR-5B                           \\
2& New Classes                     & 2        & 10     & 500           & \checkmark  &                &               & CIFAR-5A & CIFAR-50A & CIFAR-50B                          \\
3& New Ckpt                     & 2        & 10     & 500           & \checkmark  &  \checkmark    &               & CIFAR-5A & CIFAR-50A & CIFAR-50B                          \\
4& New Dataset                  & 2        & 10     & 500           & \checkmark  &                & \checkmark    & CIFAR-5A & CIFAR-50A & Tiny-ImageNet                      \\
5& Three Task                   & 3        & 30     & 300 $\times$2 & \checkmark  &                &               & CIFAR-10 & CIFAR-50A $\times$ 2 & CIFAR-50B $\times$ 2    \\
\bottomrule
\end{tabular}}
\end{small}
\caption{\textbf{Experimental setup:} Our experiments mainly focus on the sequential transfer
between two tasks, with various conditions at meta-testing time, such as unseen image classes,
unseen pretrained network and unseen data domain. We also studied sequential transfer of three tasks
in Experiment 5.}
\label{tab:tasks}
\end{center}
\vspace{-0.15in}
\end{table*}

\vspace{-0.1in}
\subsection{Implementation details}
\label{sec:impl}
A separate three-layer stacked LSTM meta-learner is learned for each layer of the student network
(except the classification layer). The meta-learner uses ReLU activation functions in the hidden
layers and $\tanh$ in the output layer. The weights and biases of the output layer are initialized
to 0 and 1 respectively, to produce a reasonable value at the starting time. A learnable scaling
coefficient is then applied to the output to adjust the range. For the classification layer, we
apply standard SGD without a meta-learner.
\vspace{-0.1in}
\paragraph{Meta-learner specification:}
For convolutional layers, we take the average over the spatial window of the pre- and post-
activations. The convolutional kernel of size $k_H \times k_W \times C_{in}$ is flattened to a
vector as the input to the meta-learner, which outputs the weight updates of size $k_H \times k_W
\times C_{in}$. For example, for a $3\times3\times10$ convolutional layer, the input dimension of
the meta-learner is $3 \times 3 \times 10$ (weights) $+ 3 \times 3 \times 10$ (gradients) $ +10$
(pre-act) + 1 (post-act) $= 191$, and the output dimension is $3
\times 3 \times 10=90$. For other layers, $k_H \times k_W$ is not applicable, and the input is
$C_{in} \times 3$ (weight, gradient, pre-act) $+1$ (post-act) and the output is $C_{in}$.

\vspace{-0.1in}
\paragraph{Baselines:} We compare our proposed method to several baselines:
\vspace{-0.1in}
\begin{itemize}[leftmargin=*]
	\item \textbf{SGD} performs standard SGD on new tasks with the same learning rate as Task A.
\vspace{-0.05in}
    \item \textbf{SGD $\times$ 0.1} is standard SGD with 0.1$\times$ learning rate. This is to see
    whether forgetting on Task A can be traded off with learning progress on Task B.
\vspace{-0.05in}
    \item \textbf{Learning without Forgetting (LwF)}~\citep{lwf} distills new data on the old
    classification branch as additional regularization. We validate the regularization coefficient
    and set it to 1.0.
\vspace{-0.05in}
    \item \textbf{Elastic Weight Consolidation (EWC)}~\citep{ewc} adds a quadratic regularizer on
    the weights, where the regularization strength is computed as a diagonal approximation of the
    Fisher information matrix. We validate the regularization coefficient and set it to 1.0.
\vspace{-0.05in}
	\item \textbf{Hard Attention to Task (HAT)}~\citep{hardattend} produces a task-specific hard
	attention on the hidden units of each layer, which helps allocate a fixed portion of the network
	towards each task. We tune the hyperparameters and set $s_{\max}$ to 3.0 and $\lambda$ to 0.75.
\vspace{-0.05in}
	\item \textbf{Model-Agnostic Meta-Learning (MAML)}~\citep{maml} 
	unrolls the readout SGD steps for 10 steps on Task B, and backprop through SGD from the
	``query'' set to learn a good representation of the backbone. During meta-test we follow the
	same protocol as \textbf{SGD} and use linear readout to get accuracy of old/new task. Note that
	this is an attempt to replicate the MAML method proposed in ~\citep{oml} in our experimental
	settings.
\vspace{-0.05in}
	\item \textbf{Representation Learning (Rep)} is a representation learning baseline on the
	meta-training set. 
	We use the meta-training 50 classes plus data from Task A to per-train a representation backbone
	using cross entropy classification objective. During meta-test we follow the same protocol as
	\textbf{SGD}.
\end{itemize}


\begin{figure*}
\vspace{-0.05in}
\begin{center}
\includegraphics[height=0.85\linewidth, trim={10.6cm 1cm 10.6cm 0.8cm},clip, angle=270]{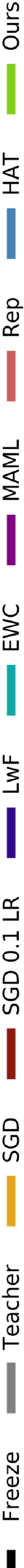}
\includegraphics[width=0.48\linewidth, trim={0cm 0cm 0cm 0cm},clip]{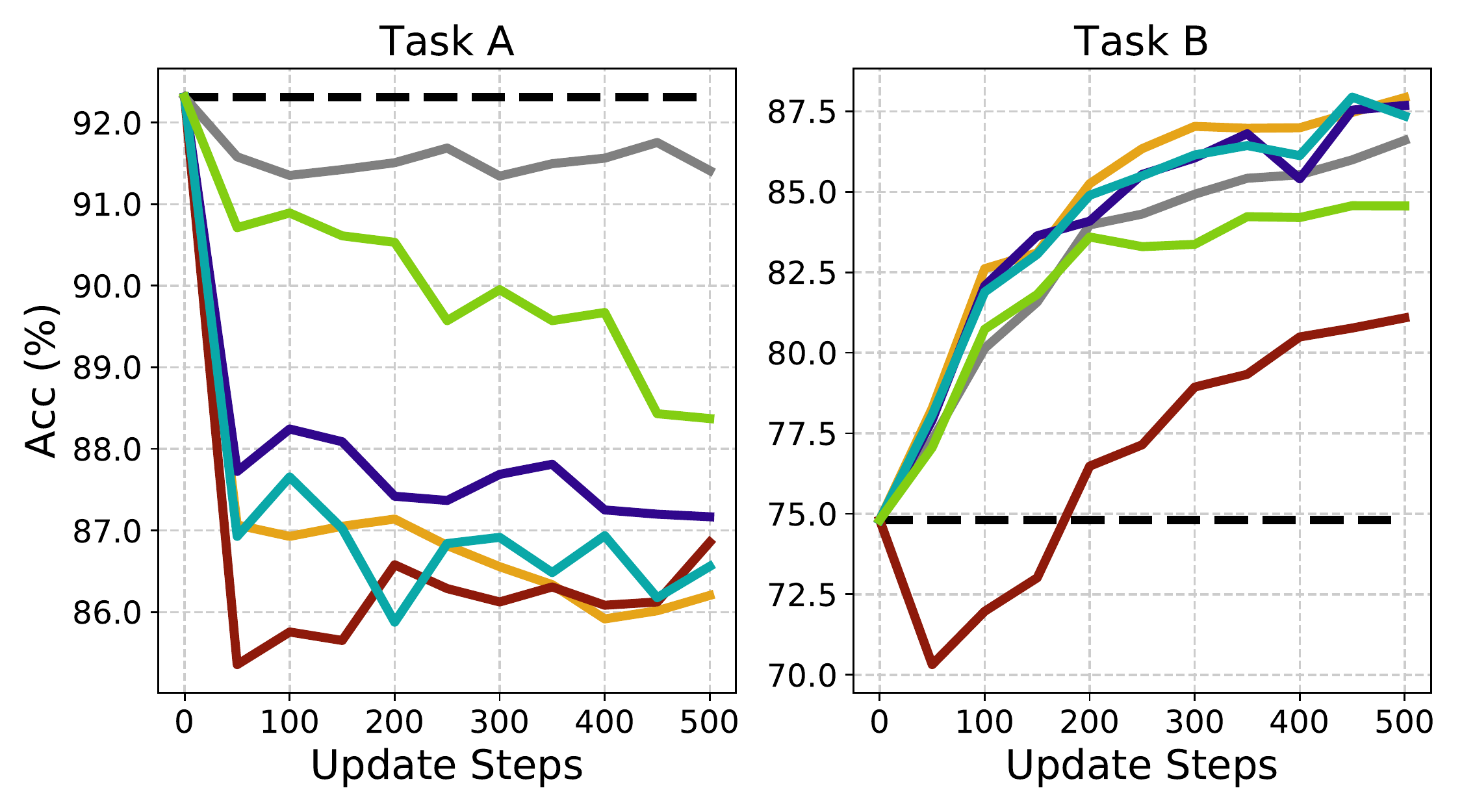}
\includegraphics[width=0.48\linewidth, trim={0cm 0cm 0cm 0cm},clip]{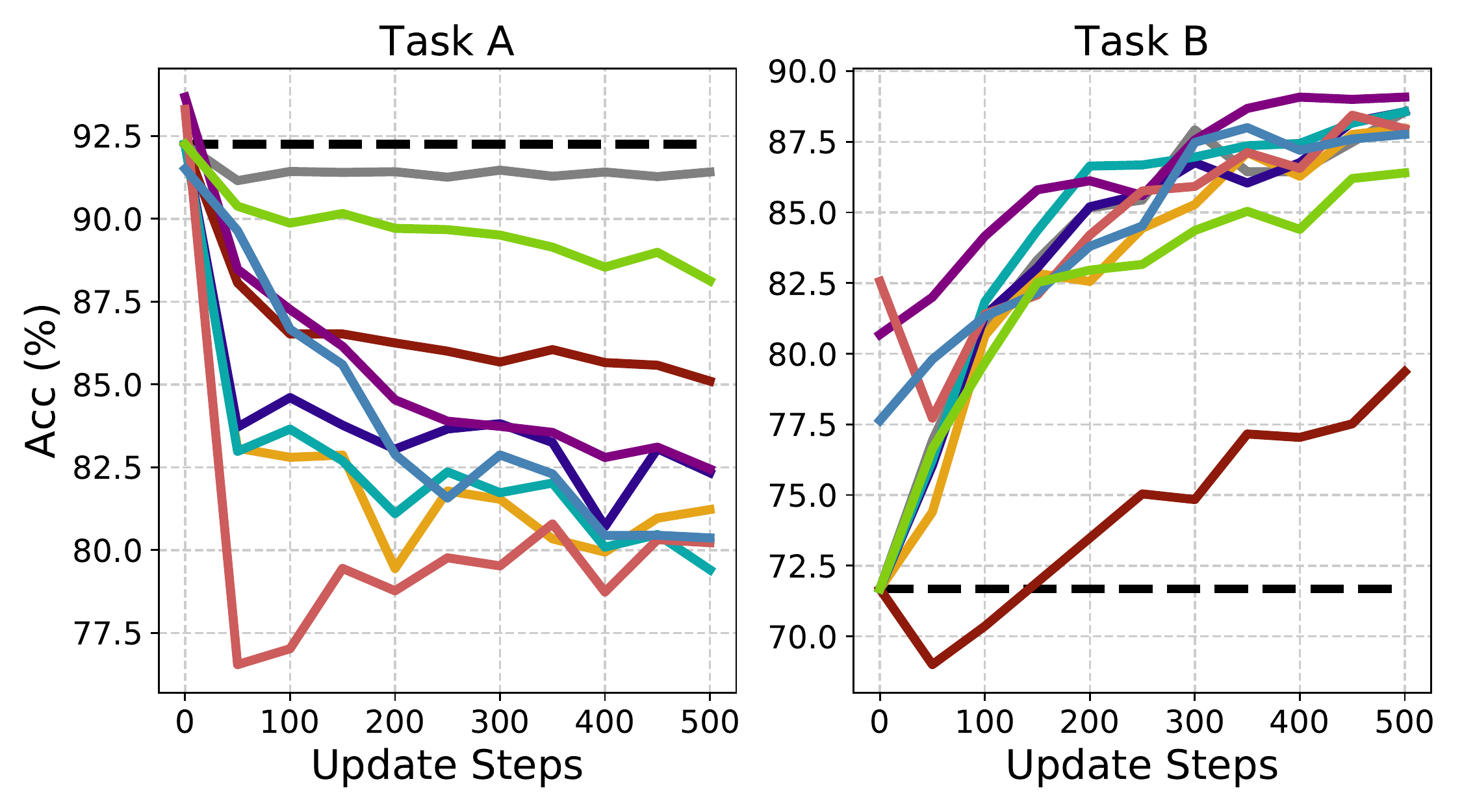}
\vspace{-0.2in}
\caption{
\textbf{Left:} Exp 1: CIFAR 5A $\mapsto$ 5B.
\textbf{Right:} Exp 2: CIFAR 5A $\mapsto$ 100 with unseen classes. 
}
\label{fig:exp1}
\end{center}
\vspace{-0.15in}
\end{figure*}

\begin{figure*}[t]
\begin{center}
\includegraphics[height=0.85\linewidth, trim={10.6cm 1cm 10.6cm 0.8cm},clip, angle=270]{figures/legend.pdf}
\includegraphics[width=0.48\linewidth]{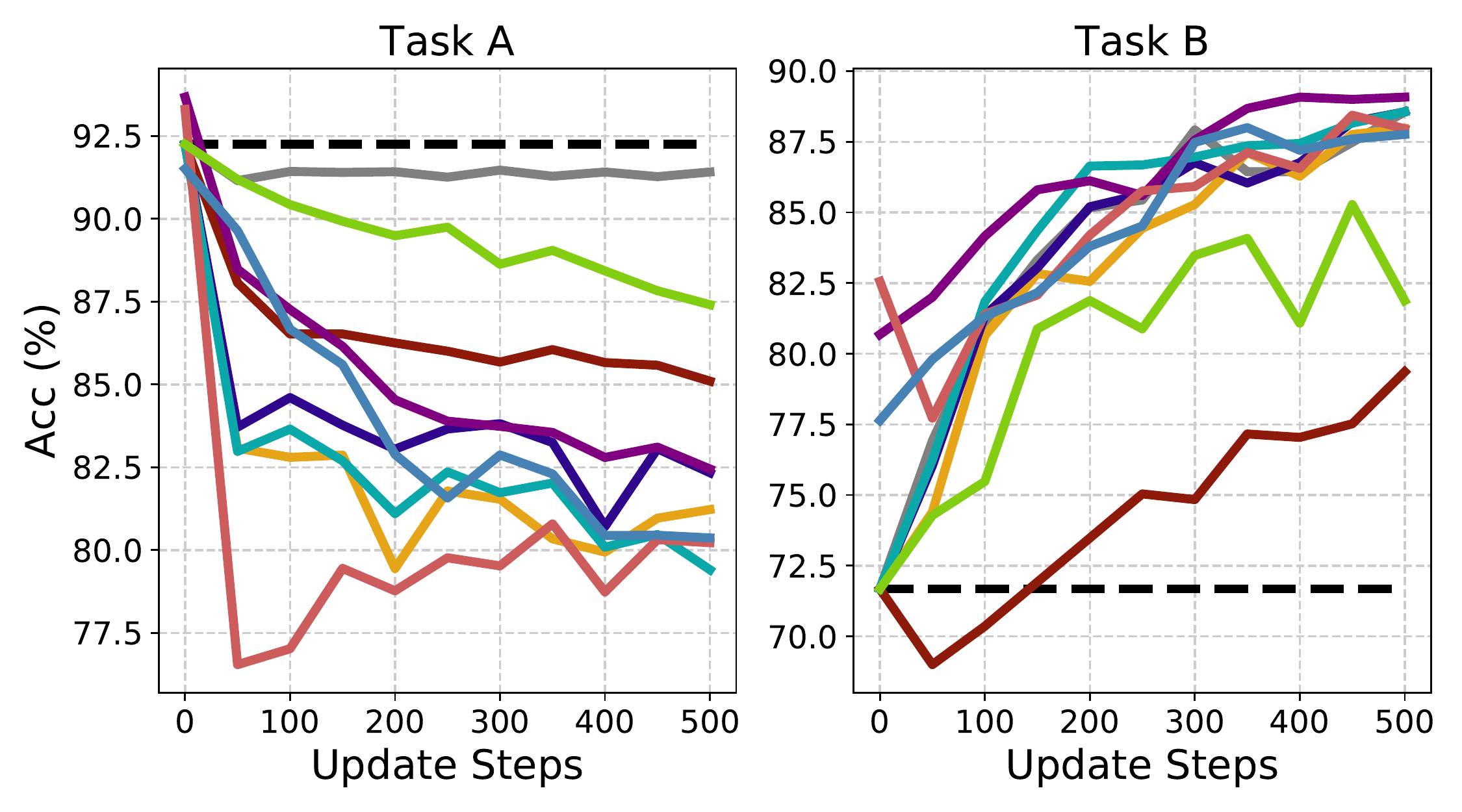}
\includegraphics[width=0.48\linewidth]{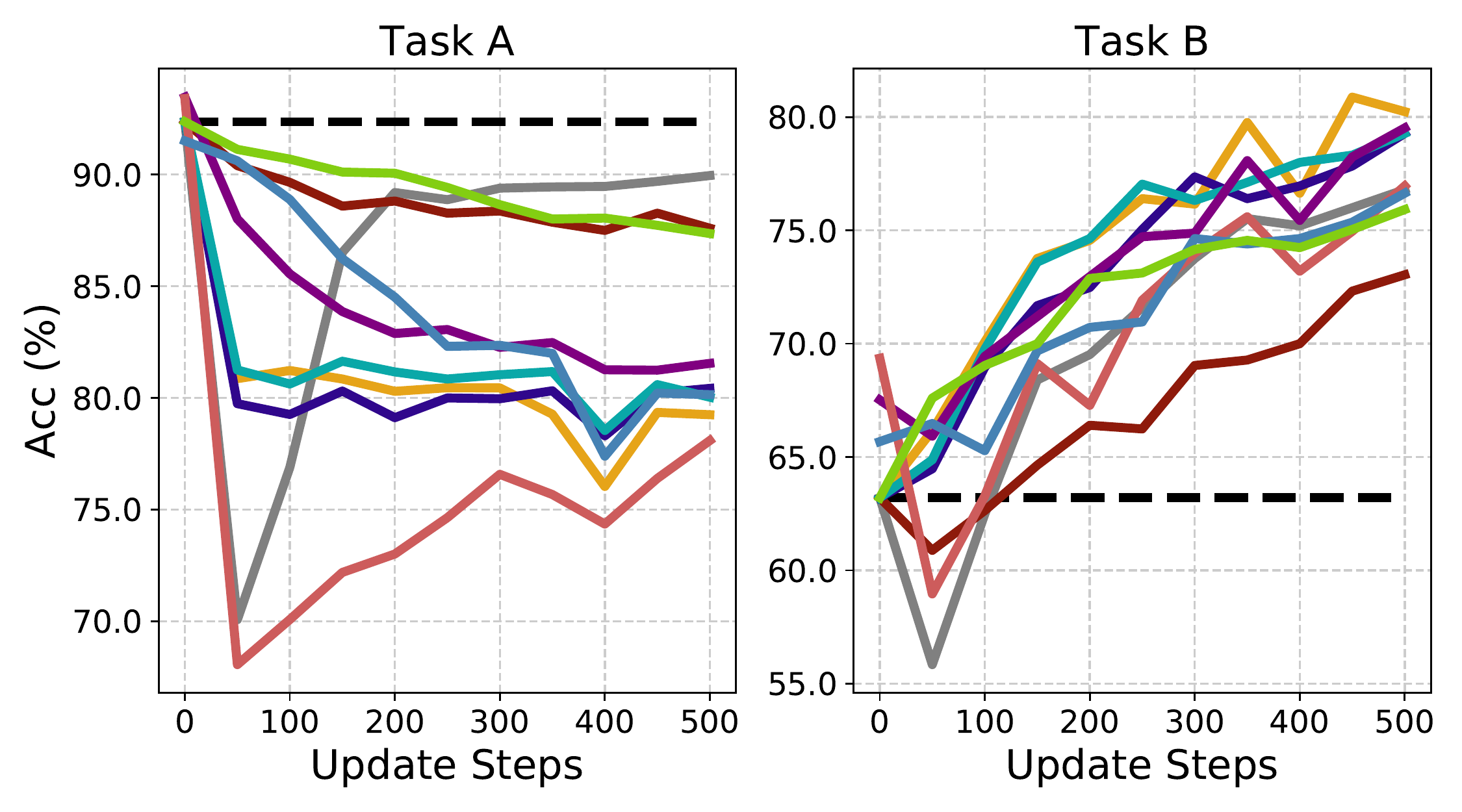}
\vspace{-0.2in}
\caption{\textbf{Left:} 
Exp 3: CIFAR 5A $\mapsto$ 100 with an unseen init checkpoint. 
\textbf{Right:} Exp 4: CIFAR5A $\mapsto$ Tiny-ImageNet (unseen dataset). 
}
\label{fig:exp3}
\end{center}
\vspace{-0.15in}
\end{figure*}

\vspace{-0.1in}
\subsection{Experiment 1: Sequential learning on two tasks}
We first conduct experiments on CIFAR-10 to verify the effectiveness of our meta-learner.
To ensure that the meta-learner does not overfit to the training examples, we split the data of Task
B into two parts evenly, denoted as $\mathcal{D}_{B_1}$ and $\mathcal{D}_{B_2}$ respectively. At
meta-training time, only $\mathcal{D}_{B_1}$ will be used; at meta-test time, only
$\mathcal{D}_{B_2}$ will be used.


\vspace{-0.1in}
\paragraph{CIFAR-10:} We split the CIFAR-10 dataset into two subsets, the first subset (CIFAR-5A)
consists of the first 5 classes (``airplane'', ``automobile'', ``bird'', ``cat'', ``deer'') and the
second subset (CIFAR-5B) consists of the remaining 5 classes (``dog'', ``frog'', ``horse'',
``ship'', ``truck''). A ResNet-32 \citep{resnet} network is first pre-trained on CIFAR-5A, with all
BatchNorm~\citep{batchnorm} layers replaced by GroupNorm~\citep{groupnorm}, using a learning rate of
0.1 and momentum of 0.9. During meta-learning, we use learning rate 0.1 without momentum. For
pre-training we use 128 examples as a mini-batch, and for meta-learning we use 128 for the teacher
and 64 for the student (Task B only). The meta-learner is a 3-layer LSTM with 64 hidden units for
the convolutional kernels and 32 hidden units for $\gamma$ or $\beta$ of the GroupNorm layers. We
set the curriculum threshold to 20 and T-BPTT step to 5 initially and increase them by 5 and 2 every
300 episodes until 30 and 9, respectively. The loss scaling coefficient is set to 300. We train the
meta-learner using the Adam optimizer with learning rate 1e-3 for a total of 900 episodes. At
meta-test time, we take the activations before the last layer and train a linear readout network
using 500 Adam optimizer steps with a learning rate of 1e-1.

\begin{figure}[t]
\begin{center}
\vspace{-0.05in}
\includegraphics[width=0.95\columnwidth,trim={0cm 0 0cm 0},clip]{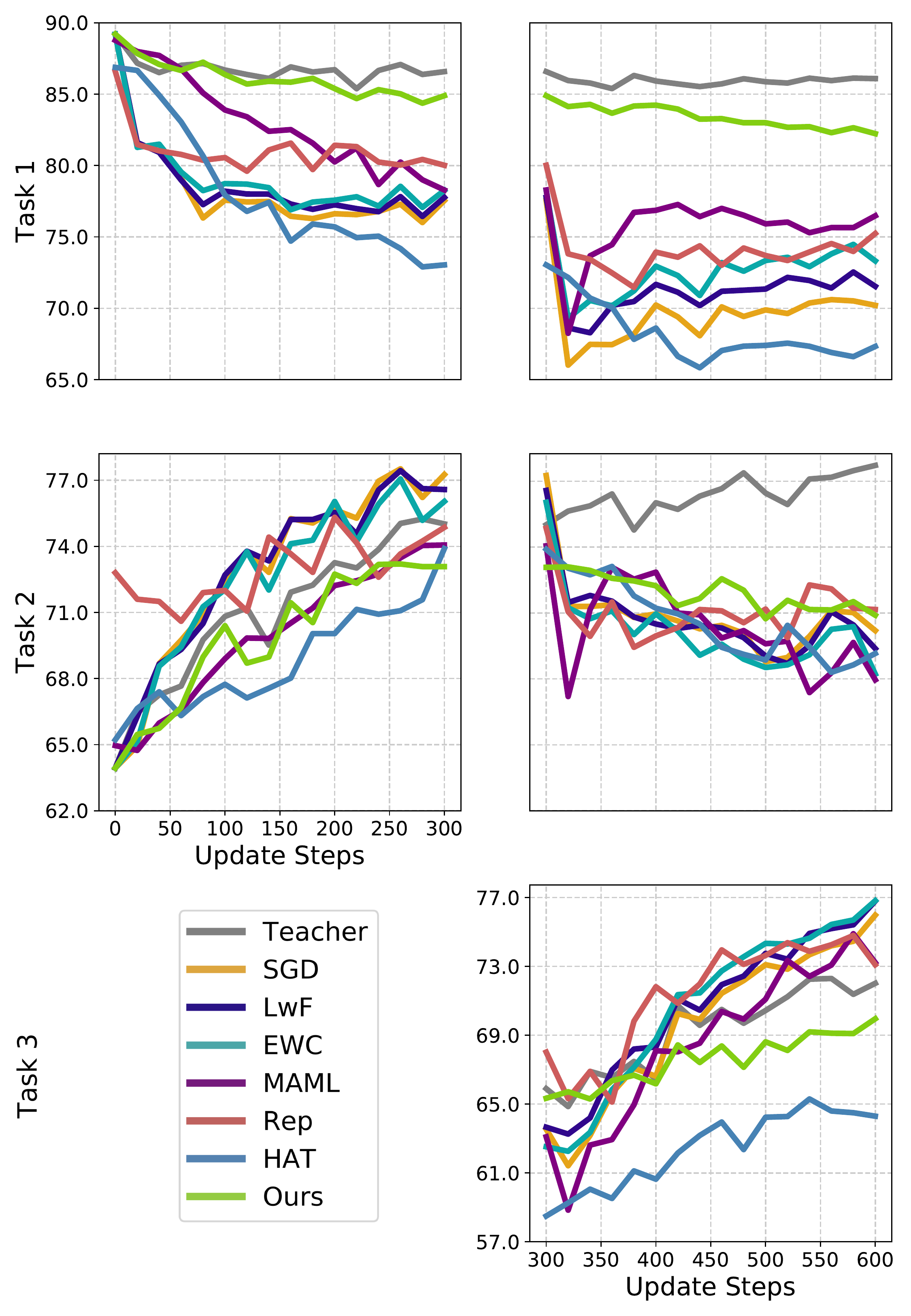}
\vspace{-0.2in}
\caption{
Exp5: CIFAR-10 $\mapsto$ 100. Three sequential tasks. Column 1 shows Task 1-2 accuracy while learning on Task 2 only; Column 2 shows Task 1-3 accuracy while learning on Task 3 only.
}
\label{fig:exp5}
\end{center}
\vspace{-0.3in}
\end{figure}

%

\vspace{-0.1in}
\paragraph{Results:} Figure~\ref{fig:exp1} shows results on CIFAR-10. The accuracy is the mean of 5 runs. All entries in the table use the proposed readout measure. We found that reducing learning rate cannot help prevent from catastrophic forgetting. Our meta-learner outperforms other methods by a large margin on Task A. On Task B, our meta-learner has very similar performance to the teacher network, which matches our expectation.

\vspace{-0.1in}
\subsection{Experiment 2: Generalizing to unseen classes}
In Experiment 1, Task B is the same for both training and testing. To verify the generalization
ability of our model to unseen classes, we utilize the CIFAR-100 dataset, to construct two different
tasks, B$_1$ and B$_2$ from disjoint subset of data. We start from an initial model trained on
CIFAR-5A, at meta-training time we train a meta-learner on Task B$_1$, while at meta-test time we
evaluate meta-learner on Task B$_2$. Unlike the previous experiment, the class definition changes
from meta-training to meta-testing. This is a more practical setting since in reality we do not know
a priori which new task the model needs to adapt to.

\begin{figure}[tbp]
\vspace{-0.05in}
\includegraphics[width=1.0\linewidth, height=4cm, trim={0cm 0cm 0cm 0cm},clip]{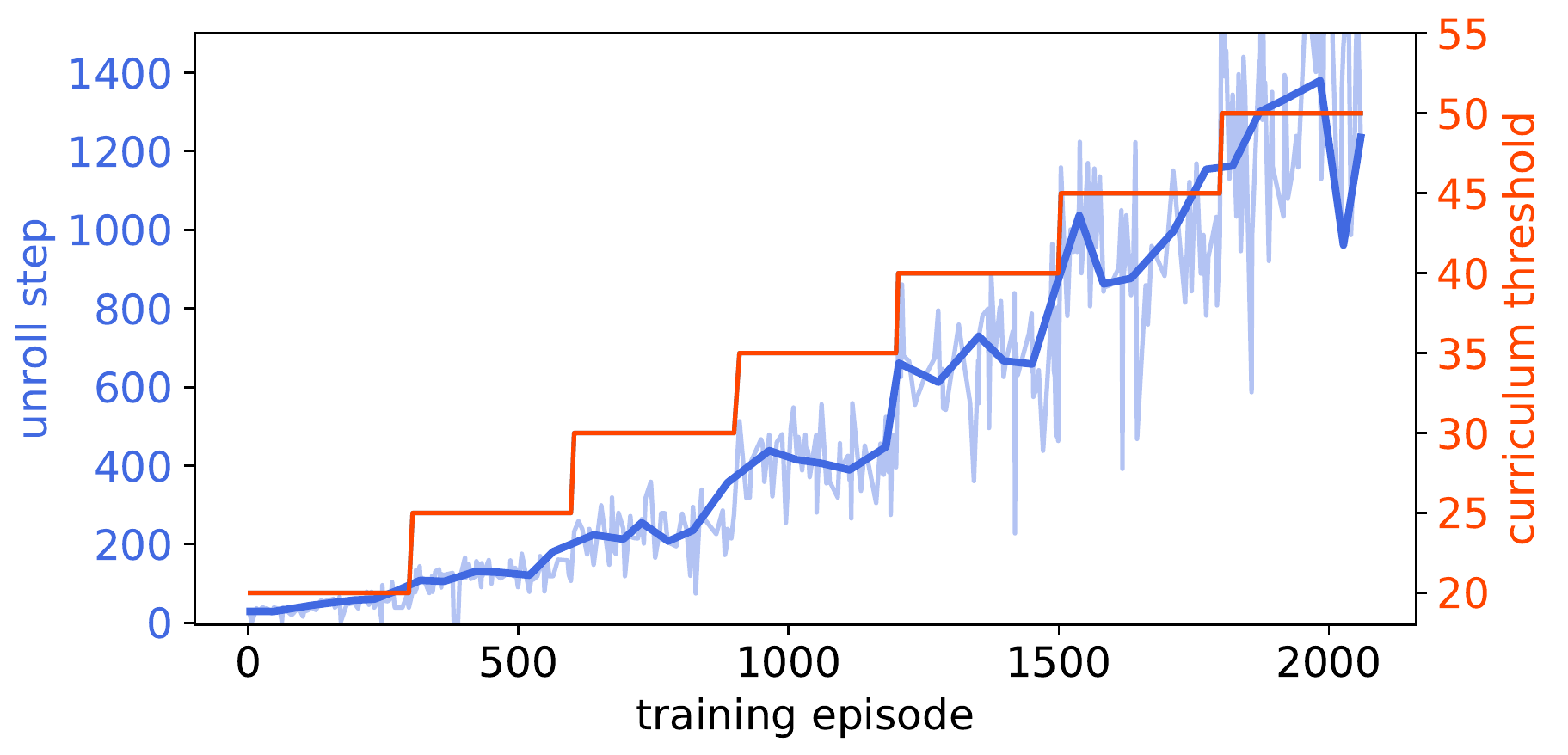}
\vspace{-0.35in}
\caption{\small Training curve on Experiment 2}
\label{fig:training curve}
\end{figure}

\vspace{-0.1in}
\paragraph{CIFAR-100:} We split CIFAR-100 dataset into two subsets. The first subset consists of the
first 50 classes, the second subset consists of the remaining 50 classes. At meta-training time we
randomly sample 5 classes from the first subset to constitute Task B$_1$ at the beginning of every
episode. And at meta-test time, we randomly sample 5 classes from the second subset to constitute
Task B$_2$, and we repeat it for 5 times to take the average performance. Task B$_1$ and Task B$_2$
have no overlap for both images and classes. We set the curriculum threshold to 20 and the T-BPTT
step to 5 initially and increase them by 5 and 2 every 300 episodes until 50 and 17 and train the
meta-learner for a total of 2k episodes. Figure~\ref{fig:training curve} illustrates the curriculum
threshold schedule and number of unrolled steps to indicate the training progress. Other
hyperparameters are kept the same as CIFAR-10 in this experiment.

\vspace{-0.16in}
\paragraph{Results:} As shown in Figure~\ref{fig:exp1}, our meta-learner generalizes well to unseen
classes and clearly outperforms other baselines. By the end of 500 update steps, the representation
forgetting on Task A is 4\% for our model, compared to over 10\% for SGD, EWC, and LwF; meanwhile on
Task B our model performs much better than SGD $\times$0.1, only $\approx $2\% behind other baselines.

\vspace{-0.1in}
\subsection{Experiment 3: Generalizing to unseen initialization}
In Experiment 3, we remove the assumption on a fixed initialization checkpoint. In order for the
meta-learner to generalize to different initialization state, during training time we provide 100
different pretrained checkpoints and each training episode uses a different checkpoint. At test
time, an unseen checkpoint is provided.

\vspace{-0.16in}
\paragraph{Results:} As shown in Figure~\ref{fig:exp3}, our meta-learner generalizes well to
unseen initialization (as well as unseen classes) and still outperforms other baselines. The
performance drops a little as expected, since the meta-learner model has to learn to adapt different
initializations during meta training/test.


\vspace{-0.1in}
\subsection{Experiment 4: Generalizing to unseen domain}
In Experiment 1, 2, and 3. All tasks are inducted from CIFAR-10/100 datasets which have similar domain. To
verify the generalization ability of our model to unseen domain, we further utilize Tiny-ImageNet as
an unseen new dataset. It has different image resolution/data distribution comparing to CIFAR, while
they both contain natural images thus the learned representation can be shared.

\vspace{-0.12in}
\paragraph{Tiny-ImageNet:} Tiny-ImageNet has 200 classes in total. At meta-training time we use the
same protocol as mentioned in Experiment 2 and keep Tiny-ImageNet data intact. At meta-test time, 5
classes from TinyImageNet are randomly sampled to serve as Task B$_2$, and we repeat it for 5 times
to take the average performance.

As shown in Figure~\ref{fig:exp3}, our meta-learner can generalize well to unseen domain, by the end of 500 update steps it outperforms other baselines by a large margin on Task A except SGD $\times$0.1; meanwhile it is close to other baselines on Task B and performs better than SGD $\times$0.1.

\vspace{-0.1in}
\subsection{Experiment 5: Sequential learning on more than two tasks}
We further extend our meta-learner to perform sequential learning on three tasks (Task 1, 2, 3). For
training efficiency, ResNet-14 is utilized as the backbone network. CIFAR-10 is used as Task 1; In
meta-training, for each rollout we sample two 10-way classification tasks from the first 50 classes
of CIFAR-100; and in meta-testing we sample from the last 50 classes. TBPTT step is set to 2
initially and the loss scaling coefficient is set to 200. We train the meta-learner for 2k episode.
All other hyperparameters are kept the same as Experiment 2. At each episode, we first start from
Task 2, when the loss reaches the corresponding curriculum threshold, we keep the hidden state of
meta-learner and student network weights, increase the curriculum threshold by 5 temporarily and
shift to Task 3 when the loss reaches the threshold again, we reset meta-learner states, network
weights and threshold, and start a new episode. At meta-test time, we unroll all models on Task 2
with fixed 300 steps, then we switch to Task 3 for another 300 steps. We record read-out accuracies for all three tasks.

\vspace{-0.15in}
\paragraph{Results:}
From Figure~\ref{fig:exp5} we can see that our meta-learner can
generalize well when training on more than two tasks. At 600 steps it is very close to the Teacher
while other baselines may drop up to 20\% on Task 1. On Task 2 it performs no worse than other
baseline and on Task 3 the gap is very close, only only $\approx $2\% behind the Teacher.
Since Task 2 has not been trained till convergence, the reduction in forgetting is not as significant as Task 1.

\vspace{-0.1in}
\subsection{Visualization of meta-learner outputs}
\label{sec:viz}

\begin{figure}[tbp]
\vspace{-0.15in}
\centering
\includegraphics[width=0.95\linewidth, height=4cm, trim={0.4cm 0.5cm 0cm 0.5cm},clip]{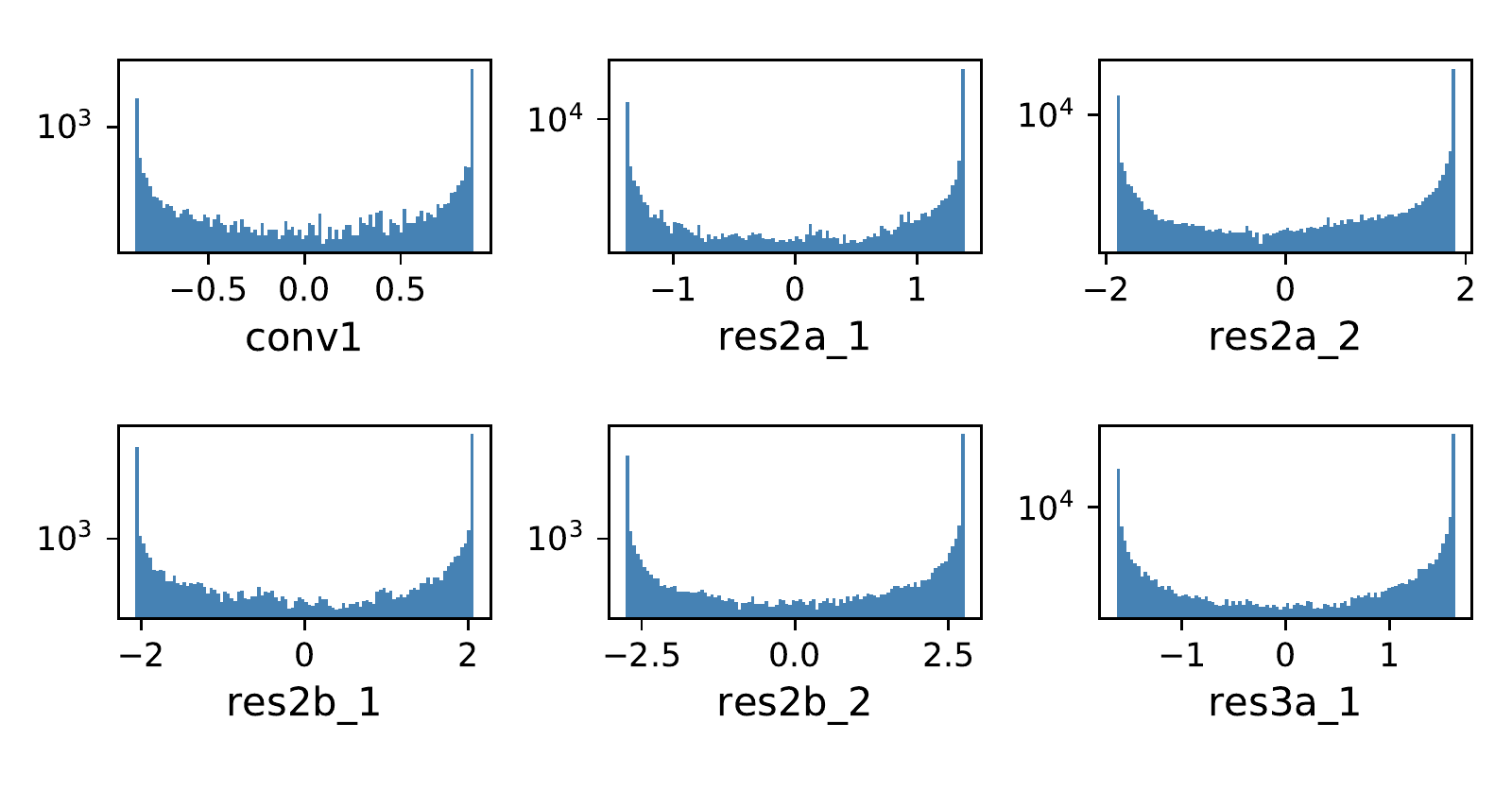}
\vspace{-0.2in}
\caption{\small Visualization of the meta-learner outputs}
\vspace{-0.15in}
\label{fig:grad_gate}
\end{figure}

To further understanding the behavior of our meta-learner, we visualize the distribution of the
output of the meta-learner (i.e. the gradient multiplier $\delta$) in Figure~\ref{fig:grad_gate}.
Our meta-learner produces non-trivial outputs that are not simply a global scaling of the learning
rate. Sometimes the multiplier can be negative, which means the final update direction is opposite
to the gradient descent direction. It shows that the meta-learner can learn to dynamically modify
the gradient direction to prevent catastrophic forgetting.
\vspace{-0.1in}
\section{Conclusion and Future Work}
Catastrophic forgetting handicaps state-of-the-art deep neural networks from learning online tasks
in the wild. This paper studies the effect of representational forgetting in a sequential learning
framework. In particular, we propose to add a linear readout layer to test the amount of forgetting
at the representation level, where a significant drop in performance on old tasks is still observed,
consistent with prior literature. We then propose to train a meta-learner to predict the weight
updates, with supervision from a multi-task teacher network. Our meta-learner is able to overcome
catastrophic forgetting while improving its performance on new tasks. We further verify that our
meta-learner has the ability to generalize to unseen classes, unseen checkpoint initializations, and unseen datasets.
Currently we have made the meta-learner successful at predicting weight updates for up to 600 steps,
but we still find it challenging to let it generalize to even longer sequences. In the future, we
expect these issues can be addressed by training the meta-learner with longer sequences and more
sequential tasks with more computational resources.

\bibliography{ref}
\bibliographystyle{icml2020}
\end{document}